\documentclass[sigconf,screen]{acmart}

\AtBeginDocument{%
  }

\acmConference[ACM-BCB '24]{15th ACM International Conference on Bioinformatics, Computational Biology and Health Informatics}{November 22, 2024}{Shenzhen, Guangdong Province, P.R. China}

\RequirePackage{graphicx}
 \usepackage{booktabs}
\usepackage[most]{tcolorbox}
\usepackage{xcolor}
\usepackage{bm}
\usepackage{graphicx}

\usepackage{longtable}
\makeatletter
\def\set@curr@file#1{\def\@curr@file{#1}} 
\makeatother
\usepackage[load-configurations=version-1]{siunitx} 

\usepackage[mathlines, switch]{lineno}

\usepackage{booktabs}

\usepackage{graphicx} 
\usepackage{booktabs} 
\usepackage{multirow} 

\title[TrialDura: Clinical Trial Duration Prediction]{TrialDura: Hierarchical Attention Transformer for Interpretable Clinical Trial Duration Prediction}

\author{Ling Yue}
\email{yuel2@rpi.edu}
\affiliation{%
  \institution{Rensselaer Polytechnic Institute}
  \city{Troy}
  \state{NY}
  \country{USA}
}

\author{Jonathan Li}
\email{lij47@rpi.edu}
\affiliation{%
  \institution{Rensselaer Polytechnic Institute}
  \city{Troy}
  \state{NY}
  \country{USA}
}

\author{Sixue Xing}
\email{xings@rpi.edu}
\affiliation{%
  \institution{Rensselaer Polytechnic Institute}
  \city{Troy}
  \state{NY}
  \country{USA}
}

\author{Md Zabirul Islam}
\email{islamm11@rpi.edu}
\affiliation{%
  \institution{Rensselaer Polytechnic Institute}
  \city{Troy}
  \state{NY}
  \country{USA}
}

\author{Bolun Xia}
\email{xiab3@rpi.edu}
\affiliation{%
  \institution{Rensselaer Polytechnic Institute}
  \city{Troy}
  \state{NY}
  \country{USA}
}

\author{Jintai Chen}
\email{cjt147@illinois.edu}
\affiliation{%
  \institution{University of Illinois Urbana-Champaign}
  \city{Chempaign}
   \state{IL}
  \country{USA}
}

\author{Tianfan Fu}
\email{fut2@rpi.edu}
\affiliation{%
 \institution{Rensselaer Polytechnic Institute}
 \city{Troy}
 \state{NY}
 \country{USA}
 }

\begin{document}

\begin{abstract}
The clinical trial process, a critical phase in drug development, is essential for developing new treatments. The primary goal of interventional clinical trials is to evaluate the safety and efficacy of drug-based treatments for specific diseases. However, these trials are often lengthy, labor-intensive, and expensive. The duration of a clinical trial significantly impacts overall costs, making efficient timeline management crucial for controlling budgets and ensuring the economic feasibility of research.
To address this issue, We propose TrialDura, a machine learning-based method that estimates the duration of clinical trials using multimodal data, including disease names, drug molecules, trial phases, and eligibility criteria. Then, we encode them into Bio-BERT embeddings specifically tuned for biomedical contexts to provide a deeper and more relevant semantic understanding of clinical trial data. Finally, the model’s hierarchical attention mechanism connects all of the embeddings to capture their interactions and predict clinical trial duration. 
Our proposed model demonstrated superior performance with a mean absolute error (MAE) of 1.04 years and a root mean square error (RMSE) of 1.39 years compared to the other models, indicating more accurate clinical trial duration prediction. 
Publicly available code can be found at: \url{https://anonymous.4open.science/r/TrialDura-F196}. 
\end{abstract}



\maketitle

\section{Introduction}




The clinical trial process, an essential step in drug development, constitutes a critical phase in bringing new medical treatments to fruition. These trials serve as the pivotal gateway for assessing the safety and efficacy of drug-based interventions in addressing various diseases prevalent in the human body~\citep{vijayananthan2008importance}. However, despite their paramount significance, clinical trials are characterized by many challenges, ranging from their time-consuming nature to their exorbitant costs and relatively low approval rates. Spanning over an extensive duration of 7 to 11 years on average, the clinical trial timeline emerges as a prominent obstacle in the drug development landscape~\citep{general_trial_time}. This protracted duration not only prolongs the time-to-market for potential treatments but also escalates the associated expenses significantly. With an average cost exceeding 2 billion dollars per trial, the financial burden imposed by these endeavors is significant. Furthermore, the low approval rates, hovering around a mere 15\%, underscore the complexity and uncertainty inherent in the clinical trial process~\citep{huang2022artificial}. 

One of the primary determinants influencing the overall cost of clinical trials is their duration. Extended trial periods necessitate more resources, more staffing, more medical supplies, and more facility usage, thus increasing operational expenses. Moreover, the prolonged data collection and analysis phases entail more financial investment in technology and personnel. Consequently, efficiently managing the timeline of clinical trials is paramount in controlling the budget and optimizing the economic feasibility of research endeavors in the pharmaceutical domain. Against this backdrop, integrating machine learning methodologies emerges as a promising avenue for enhancing the efficiency and accuracy of clinical trial duration estimation. By leveraging computational algorithms to analyze intricate datasets, machine learning facilitates the prediction of trial duration with improved precision and reliability. In this paper, we delve into the application of various machine learning techniques, including traditional methodologies such as Linear Regression (LR)~\citep{weisberg2005applied}, Gradient Boosting Decision Tree (GBDT)~\citep{ke2017lightgbm}, adaptive Boosting (AdaBoost)~\citep{ying2013advance}, Random Forest (RF)~\citep{breiman2001random}, and a cutting-edge hierarchical deep learning framework that we designed specifically for the input data, to predict the duration of clinical trials.

Our research also produces a comprehensive dataset explicitly tailored for clinical trial duration prediction. This dataset encapsulates diverse eligibility criteria of various clinical trials, including patient demographics, treatment protocols, and regulatory parameters, thereby offering a holistic perspective for predictive modeling. Through extensive experimentation and analysis, we evaluate the efficacy of seven distinct machine learning methodologies in predicting clinical trial duration. The overarching objective of this study is to demonstrate the potential of machine learning in optimizing the clinical trial process, thereby offering insights into the expected duration of clinical trials. By showing the efficacies of various predictive methods, we aim to provide insights that can inform decision-making processes within the pharmaceutical industry, ultimately fostering more efficient and cost-effective drug development practices. Through our empirical findings, we strive to pave the way for adopting data-driven approaches in the landscape of clinical trial management and accelerate the translation of scientific innovations into tangible pharmaceutical solutions. 

The main contributions can be summarized as 
\begin{enumerate}
\item Problem: We are the first to identify and propose the problem of clinical trial duration prediction, an essential issue in clinical trial planning, and formulate it into an AI-solvable problem.
\item Method: We propose multiple traditional machine learning models to address this problem and design a novel hierarchical attention mechanism, a specialized end-to-end deep learning framework tailored for predicting clinical trial duration. This framework captures multi-modal data elements within clinical trials and leverages large-scale biomedical language models (Bio-BERT).
\item Results: The proposed method (TrialDura) demonstrated superior performance, achieving an MAE of 1.044 years and an RMSE of 1.390 years. In comparison to the best baseline method, TrialDura achieves relative reductions of 9\% and 7\% in MAE and RMSE, respectively. 
\item Interpretability: Our method exhibits desirable interpretability thanks to the hierarchical attention mechanism, which would assist clinicians in making informed decisions by providing insights into how predictions were reached. 
\end{enumerate}


    

\section{Related Works}
\label{sec:related}

Earlier works in the area of machine learning for clinical trial predictions involve two major directions: predicting individual patient outcomes and predicting the probability of trial success~\citep{askin-survey}. 

In terms of predicting individual patient outcomes, the early approaches use traditional methods in machine learning. For instance, \cite{wu-genetic-lesion} utilized support vector machines (SVMs) to forecast genetic lesions based on cancer clinical trial documents. \cite{raj20evaluation} utilized gradient-boosted decision trees (GBDTs) to predict improvements in symptom scores by integrating treatment symptom scores and EEG measures in the context of antidepressant treatment for depressive symptoms. Meanwhile, \cite{hong20predicting,du2023abds} directed their efforts towards projecting clinical drug toxicity based on features related to drug properties and target properties, employing an ensemble classifier consisting of weighted least squares support vector regression. Machine learning can also play a pivotal role in generating simulated data to identify more efficient statistical outcome measures \citep{sangari-ml-for-breast-cancer}. One study proposes that employing an AI algorithm to forecast individual patient outcomes and pinpoint those likely to progress rapidly, leading to earlier trial endpoints, could result in shorter trial duration~\citep{lee-ai-for-trials}. 
Multi-omics data enables the characterization of individual patients in a fine-grained manner, which is crucial for precision medicine approaches. By understanding the genetic~\cite{lu2021cot,lu2022cot,chen2021data}, transcriptomic~\cite{lu2023deep,yi2018enhance}, and other molecular profiles of patients~\cite{chen2021data,fu2024ddn3,zhang2021ddn2}, treatments can be tailored to match individual disease mechanisms, potentially leading to more effective and personalized therapies~\citep{wu2022cosbin,lu2019integrated,lu2018multi}. 
Moreover, \cite{Berchialla1} developed a machine learning framework to conduct a heterogeneous treatment analysis on type 2 diabetes using randomized clinical trials that share similar inclusion and exclusion criteria as well as a set of common clinical patient characteristics.

In terms of predicting the probability of trial success, we observe machine learning being utilized to generate early detections and prognoses of disease, which can increase the probabilities of success in a clinical trial \citep{sangari-ml-for-breast-cancer}. Beyond that, machine learning can be employed to predict molecular features, target sensitivity, bioavailability, and toxicity~\citep{zhavoronkov-ai-drug-discovery} in order to minimize potential trial failures in later stages and can contribute to designing phase-II and phase-III trials that have a higher chance of success. \cite{qi-phase-3-trial} designed a Residual Semi-Recurrent Neural Network (RS-RNN) to forecast phase III trial outcomes based on phase II results. \cite{lo-2019-ml-drug-approvals} explored different imputation techniques and a range of traditional machine learning models, including logistic regression, random forest, and SVM, to forecast drug approval within 15 disease categories. Similarly, \cite{siah-2021-novartis-ai} assessed various traditional machine learning models for predicting clinical trial outcomes. In more recent works, \cite{fu2023automated,fu2022hint,lu2024uncertainty} predicted outcomes across all clinical trial phases, utilizing comprehensive trial features such as drug molecules and trial eligibility criteria and integrated multimodal data sources. 
\cite{wang2024twin} design a TWIN-GPT model to synthesize patient electronic records to simulate clinical trials. 
\cite{Kavalci2023} also introduced an improvement to the predictive model for clinical trial termination by incorporating eligibility criteria search capabilities derived from unstructured text and disease categorization characteristics into the existing study features and showed a statistically significant improvement in predicting early termination of clinical trials. Moreover, \cite{https://doi.org/10.1002/cpt.3008} created a platform that utilizes transformer technology and generative AI, incorporating multimodal data such as omics, text, clinical trial design, and small-molecule properties to forecast the results of phase II clinical trials. 
\citep{chen2024trialbench} design a data platform that contains more than 20 AI-ready clinical trial datasets, e.g., trial duration prediction, trial eligibility criteria design, trial adverse event prediction, etc. 

The earlier works in machine learning for clinical trial predictions primarily focus on two aspects: predicting individual patient outcomes and predicting the probability of trial success. However, these approaches, while innovative and beneficial in their respective domains, did not directly address the issue of clinical trial duration. Our approach, however, seeks to fill this gap by specifically targeting the trial duration problem. We cannot simply adapt the methodologies from prior works because our task encompasses a broader scope. Our model integrates Bio-BERT embeddings, specifically tuned for biomedical contexts, providing a deeper and more relevant semantic understanding of clinical trial data. Additionally, the hierarchical attention mechanism in our model enables a nuanced analysis of relationships between various data elements, a feature not commonly addressed in previous studies.

\section{Methodology}
\label{sec:method}

\subsection{Formulation of Clinical Trial Duration
Prediction}
\label{sec:formulation}

A {\textit{clinical trial}} is an organized endeavor to evaluate the safety and efficacy of a {\it treatment set} aimed at combating a {\it target disease set}, according to the guidelines laid out in the {\it trial eligibility criteria}, for a select group of patients.

\begin{definition}[\textbf{Clinical Trial Phases}]
The phases of clinical trials signify distinct stages in the trial process, each with a specific focus and set of objectives. These phases are:
\begin{itemize}
\item Phase 1: Initial evaluation of the new drug's safety and profiling of its pharmacological profile.
\item Phase 2: Assessment of the drug's efficacy and side effects in a larger patient group.
\item Phase 3: Confirmation of the drug's effectiveness, monitoring of side effects, comparison to commonly used treatments, and collection of information that will allow the drug to be used safely.
\item Phase 4: Post-marketing studies delineate additional information, including the drug's risks, benefits, and optimal use.
\end{itemize}
The phase information is denoted $\mathcal{P}$, which is a one-hot vector.
\end{definition}

\begin{definition}[\textbf{Treatment Set}]
The treatment set encompasses a variety of drug candidates, 
denoted as \underline{$\mathcal{M} = \{ m_1, \cdots, m_{K_{m}} \}$}, 
where $m_1, \cdots, m_{K_{m}}$ are $K_{m}$ drug molecules engaged in this trial. 
This study concentrates on trials to identify new applications for these drug candidates, while trials focusing on non-drug interventions like surgery or device applications are considered outside the scope of this research.
\end{definition}

\begin{definition}[\textbf{Target Disease Set}]
\label{def:disease}
For a trial addressing $K_{\delta}$ diseases, the Target Disease Set is represented by \underline{$\mathcal{D} = \{\delta_1, \cdots, \delta_{K_{\delta}}\}$}, with each $\delta_i$ symbolizing the diagnostic code \footnote{In this paper, we use ICD10 codes (International Classification of Diseases)~\citep{anker2016welcome}} for the $i$-th disease.
\end{definition}

\begin{definition}[\textbf{Trial Eligibility Criteria}]
The trial eligibility criteria encompass both inclusion (+) and exclusion (-) criteria, which respectively outline the desired and undesirable attributes of potential participants. These criteria provide details on various key parameters such as age, gender, medical background, the status of the target disease, and the present health condition.
\begin{equation}
\label{eqn:protocol_advanced}
\quad \mathcal{E} = [\bm{\pi}_{1}^{+}, ..., \bm{\pi}_{Q}^{+}, \bm{\pi}_{1}^{-}, ..., \bm{\pi}_{R}^{-}],\ \ \ \ \ \ \bm{\pi}_{k}^{+/-} \ \text{is a criterion}, 
\end{equation}
where $Q$ ($R$) is the number of inclusion (exclusion) criteria in the trial. The term $\bm{\pi}_{k}^{+}$ ($\bm{\pi}_{k}^{-}$) designates the $k$-th inclusion (exclusion) criterion within the eligibility criteria. Each criterion $\bm{\pi}$ is a sentence in unstructured natural language. 
\end{definition}

\begin{definition}[\textbf{Clinical Trial Duration}]
The duration of a clinical trial refers to the number of years the trial lasts. It is represented as a continuous value \( y > 0 \). 
\end{definition}

\noindent\textbf{Problem (Clinical Trial Duration Prediction).}
The estimation of \( y \), represented as \( \widehat{y} \), can be formulated through the machine learning model \( f_{\Theta} \), such that \( \widehat{y} = f_{\Theta}(\mathcal{P}, \mathcal{M}, \mathcal{D}, \mathcal{E}) \), where \( \widehat{y} > 0 \) denotes the estimated duration of a trial; $\Theta$ denotes the learnable parameter. In this context, \( \mathcal{P} \), \( \mathcal{M} \), $\mathcal{D}$, and \( \mathcal{E} \) refer to the phase information, treatment set, the target disease set, and the trial eligibility criteria, respectively.

For ease of exposition, Table~\ref{table:trial_example} shows an example of a real clinical trial and all the related features.

\begin{table}[ht]
\centering
\caption{A real example of a clinical trial record.}
\begin{tabular}{c|p{0.63\columnwidth}}
\toprule
Feature & Descriptions \\ 
\midrule
NCTID & NCT00610792 \\ 
disease & Ovarian Cancer \\ 
phase & II \\ 
title &  Phase 2 Study of Twice Weekly VELCADE and CAELYX in Patients With Ovarian Cancer Failing Platinum Containing Regimens \\ 
summary & This is a Phase 2, multicenter open-label, uncontrolled 2-step design.  Patients will be arranged in two groups based on the response to their last platinum containing therapy. \\
& The two groups are, 1) Platinum-Resistant Patients: patients with the progressive disease while on platinum-containing therapy or stable disease after at least 4 cycles; patients relapsing following an objective response while still receiving treatment; 
patients relapsing after an objective response within 6 months from the discontinuation of the last chemotherapy 
 and 2) Platinum-Sensitive Patients: patients who relapsed following an objective response \\ 
study type & interventional \\ 
Inclusion Criteria & ECOG performance status grade 0 or 1
; Age $\geq$ 18 and $\leq$ 75 yrs; Life expectancy of at least 3 months; LVEF must be within normal limits; ...   
\\ 
Exclusion Criteria & Chemotherapy, hormonal, radiation or immunotherapy or participation in any investigational drug study within 4 weeks of study entry; 
Pre-existing peripheral neuropathy $>$ Grade 1; 
Presence of cirrhosis or active or chronic hepatitis; ... 
Pregnancy or lactation or unwillingness to use adequate method of birth control; 
Active infection; 
Known history of allergy to mannitol, boron or liposomally formulated drugs. 
\\ 
drug &  bortezomib and pegylated liposomal doxorubicin \\ 
start date & July 2006 \\ 
completed date & September 2009 \\ 
duration & 3.2 years \\ 
sponsor & Millennium Pharmaceuticals, Inc. \\ 
outcome & withdrawn \\ 
\bottomrule
\end{tabular}
\label{table:trial_example}
\end{table}

\subsection{Broad Impact}

Accurately forecasting the duration of clinical trials offers significant benefits for trial management. By predicting trial duration, resource allocation such as staffing, budget, and facilities can be optimized, ensuring resources are available when needed to prevent inefficiencies and bottlenecks \citep{Kerali2018}. This capability is crucial for planning financial aspects, enhancing the accuracy of budgets, and ensuring efficient use of capital \citep{Baskin2019}.

Understanding trial timelines aids in devising effective participant recruitment and retention strategies, which are vital in maintaining engagement and compliance in longer trials. This strategic approach minimizes dropout rates and improves trial outcomes. Accurate forecasts also enable precise budgeting for personnel, equipment, and operational expenses, thus managing costs more effectively and reducing waste \citep{Prasad2024}.

Further, forecasting facilitates improved communication with stakeholders such as sponsors, regulatory bodies, and participants, setting realistic expectations and fostering trust, which leads to more effective collaborations \citep{Yu2024}. It also aids in securing funding by providing reliable financial plans to sponsors and grant committees.

Recognizing potential delays and their impact allows for proactive risk management, including development of contingency plans for recruitment challenges, regulatory delays, or unforeseen events. This proactive approach is supported by retrospective analyses of past trials and potentially enhanced by prospective data collection during ongoing trials \citep{Knirsch2012}.

For pharmaceutical companies, accurate forecasting of trial duration is essential for strategic planning, including market entry and product launch strategies. It determines the timing of drug approval and market availability, crucial for competitive positioning and financial planning \citep{DiMasi2016}. Forecasting also supports planning for regulatory submissions, leading to smoother interactions with regulatory bodies and a more efficient approval process, ultimately speeding up market access for new treatments \citep{ALSULTAN20201217}.

In summary, the ability to forecast the duration of clinical trials provides strategic advantages in managing the complexity, costs, and risks associated with clinical research. It enhances the efficiency and effectiveness of trial execution and plays a critical role in advancing medical innovation by bringing new treatments to market more reliably and swiftly.

\subsection{Development of Network}

Our novel approach consists of three primary phases that form an end-to-end deep learning framework. These consist of embedding generation, hierarchical attention mechanism, and neural network training. We construct a hierarchical interaction graph $G$ to connect all input data sources affecting clinical trial duration $\widehat{y}$. The interaction graph $G$ is constructed to reflect the real-world trial development process and consists of four inputs including drugs, diseases, phase, and eligibility criteria with node features of input embedding  $I_m, I_d, I_p, I_e \in \mathbb{R}^d$. The phase of each clinical trial, represented categorically from 1 to 4, is converted into a one-hot encoded vector. ${I_p}$ represent the one-hot encoded phase of a clinical trial:

\begin{equation}
\label{eqn:phase }
\quad{I_p} =
\begin{cases}
[1, 0, 0, 0]^\top, & \text{if } \text{$Phase$} = 1, \\
[0, 1, 0, 0]^\top, & \text{if } \text{$Phase$} = 2, \\
[0, 0, 1, 0]^\top, & \text{if } \text{$Phase$} = 3, \\
[0, 0, 0, 1]^\top, & \text{if } \text{$Phase$} = 4. 
\end{cases}
\end{equation}

Formally, we represent drug molecules graphs with node features of input embedding  as $\mathcal{M} = \{m_1, \cdots, m_{K_{m}}\}$ (Equation 1), disease codes as $\mathcal{D} = \{\delta_1, \cdots, \delta_{K_{\delta}}\}$ (Equation 2), and eligibility criteria as $\mathcal{E} = [\bm{\pi}_{1}^{+}, \ldots, \bm{\pi}_{Q}^{+}, \bm{\pi}_{1}^{-}, \ldots, \bm{\pi}_{R}^{-}]$, where $\bm{\pi}_{k}^{+/-}$ is a criterion (Equation 3) as follows: 
\begin{equation}
\label{eqn:Molecule Embedding}
\text{Molecule Embedding,} \quad I_{m} = \frac{1}{K_{m}} \sum_{{j}=1}^{K_{m}} f_{m}(m_{j}), \quad I_{m} \in \mathbb{R}^d \quad 
\end{equation}

We employ the Bio-BERT transformer model, a domain-specific language representation model trained on biomedical corpora, to embed the drug names where $ f_{m}$ is the molecule embedding function. The embedding for each token generated by Bio-BERT are averaged to produce a single 768-dimensional vector representing the drug.

\begin{equation}
\label{eqn:Disease Embedding}
\text{Disease Embedding,} \quad I_d = \frac{1}{K_{\delta}} \sum_{{i}=1}^{K_{\delta}} f_{\delta}(\delta_i), \quad I_d \in \mathbb{R}^d, 
\end{equation}
where $ f_{\delta}$ represents an embedding of disease using Bio-BERT, similar to the drug name embedding. The mean of these embedding provides a comprehensive 768-dimensional vector representing the disease.

\begin{equation}
\label{eqn:sentence_embedding}
\text{Word Embedding,} \quad I_s = \text{BioBERT}_{\text{CLS}}(\pi), \quad I_s \in \mathbb{R}^d
\end{equation}

where $I_s$ denotes the sentence embeddings derived from the CLS token outputs of Bio-BERT. The symbol $\pi$ represents individual sentences within the eligibility criteria. 

\begin{equation}
\label{eqn:paragraph_embedding}
\text{Sentence Embedding,} \quad I_g = f_{\text{transformer}}(I_s), \quad I_g \in \mathbb{R}^d
\end{equation}

The function $f_{\text{transformer}}$ encapsulates the transformer model's mechanism, which processes these sentence embeddings $I_s$. This function enhances inter-sentence relationships, culminating in the generation of the paragraph embedding $I_p$, which integrates the comprehensive context of the eligibility criteria.

The eligibility criteria text was initially processed using n-word embeddings from the CLS token of Bio-BERT to generate sentence embeddings. 
To create paragraph embeddings, a transformer with an attention mechanism was used to enhance sentence-to-sentence relations, which is essential for capturing the comprehensive context of the eligibility criteria that often include multiple independent rules.
The procedure involved padding sentences to 32 tokens, concatenating the inclusion and exclusion criteria embeddings to form a 64 × 768 matrix, and prefixing with a CLS token embedding to obtain a 65 × 768 matrix. After processing the inputs, the embeddings representing the drug, disease, eligibility criteria (inclusion and exclusion), and trial phases were concatenated to form a unified feature vector. The dimensions of the concatenated vector were 4 × 768, with the addition of four. The concatenation of input embeddings as an intermediate step:
\begin{equation}
\label{eqn:concat_embedding}
 \quad I_{\text{concat}} = \text{CONCAT}(I_p, I_m, I_d, I_g), \ \ \quad I_{\text{concat}} \in \mathbb{R}^d. 
\end{equation}

This comprehensive feature vector was then fed into a multilayer perceptron (MLP) for regression analysis.
The estimated trial duration prediction using the concatenated embedding as input to the MLP:
\begin{equation}
\label{eqn:Trial Duration}
\text{Trial Duration,} \ \  \widehat{y} = \text{MLP}(I_{\text{concat}}). 
\end{equation}

\begin{figure*}[ht]
\centering
\includegraphics[width=0.9\textwidth,keepaspectratio]{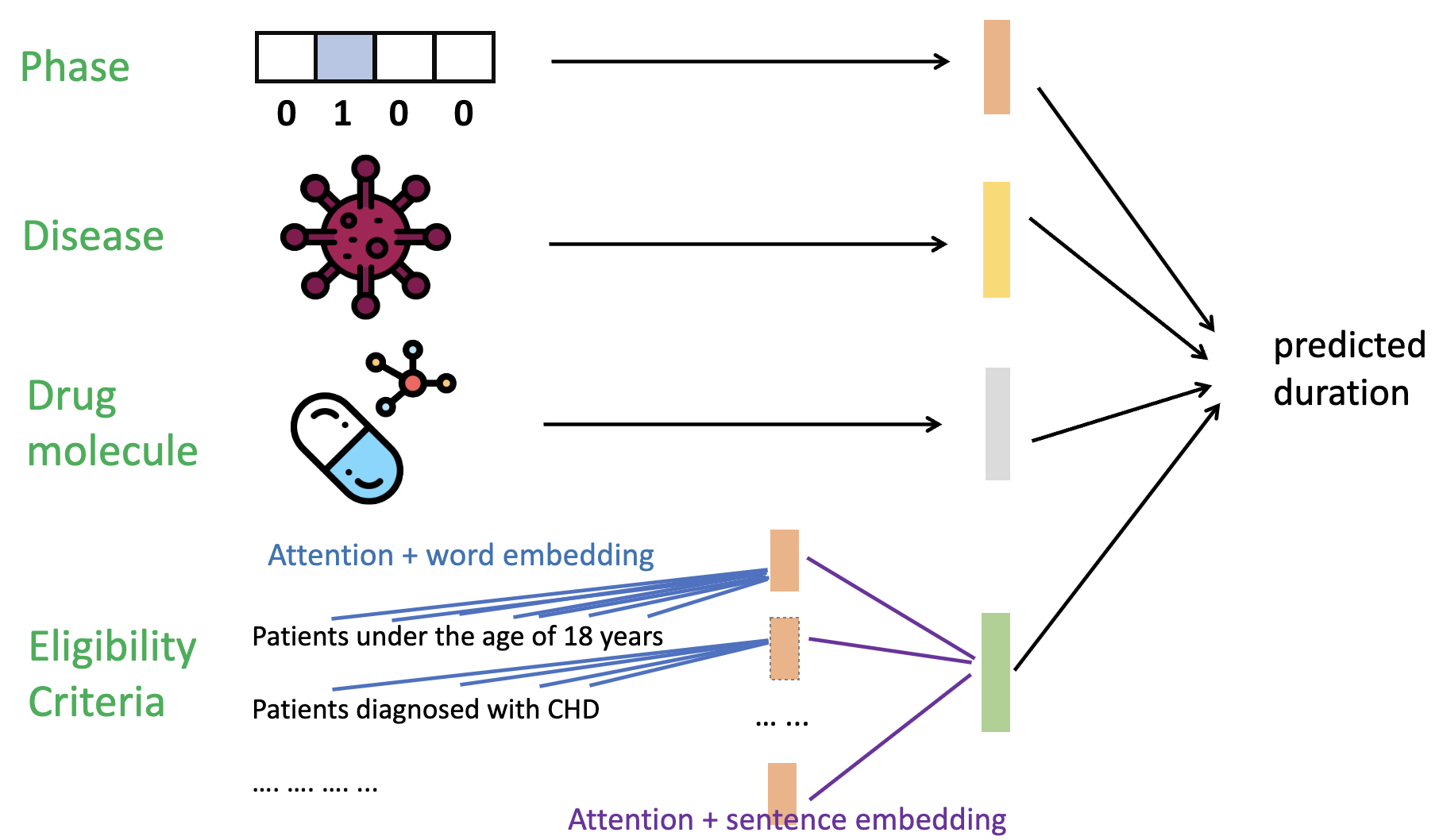}
\caption{Overview of our model. Our model takes phase information, disease, drug molecule and eligibility criteria as the input and predicts the trial duration. We design a hierarchical attention mechanism to model semantic features from eligibility criteria. The lower-level attention mechanism aggregates word embeddings while higher-level attention gathers sentence embeddings. }
\label{fig:method}
\end{figure*}

We employed the Mean Squared Error (MSE) loss function to quantify the discrepancy between predicted and actual trial duration. 
The objective function is defined by the following equations: 
\begin{equation}
\text{MSE} = {\frac{1}{n}\sum_{i=1}^{n}(y_i - \widehat{y}_i)^2},
\end{equation}
where $y_i$ and $\widehat{y}_i$ represent the true and predicted duration for the $i^{th}$ clinical trial, respectively, and $n$ is the total number of trials in the training dataset.
The Adam optimizer~\citep{kingma2014adam} was utilized to minimize the objective function and update the network weights iteratively based on the training data. The architecture of the whole model is illustrated in Figure~\ref{fig:method}. 

\section{Experiment} 
\label{sec:experiment}

\subsection{Dataset Setting}

We collected a dataset from \url{https://clinicaltrials.gov/}, a comprehensive global database of clinical trials, to determine the duration of these trials. We only consider trials that last less than 10 years, which account for 98\% of all the trials, because some extremely long values significantly affect model performance.
Each trial was represented as an XML file, from which we extracted several pieces of information, including the NCT ID, disease name, drug molecule, trial phase, time duration, and eligibility criteria. After extracting the data, we engaged in data processing to prepare the data for machine learning applications. 
The statistics for duration are summarized in Table~\ref{tab:dura_stats}. 

\begin{table*}[htbp]
\centering
\caption{Statistics for Duration (Unit: Year)}
\begin{tabular}{cccccc}
\toprule
Phase & Average & Minimum & 1st Quartile & Median & 3rd Quartile \\ \midrule
All Phases & 2.14 & 0.003 & 0.83 & 1.73 & 3.00 \\ 
Phase 1 & 1.52 & 0.003 & 0.25 & 0.87 & 2.26 \\ 
Phase 2 & 2.47 & 0.003 & 1.09 & 2.00 & 3.34 \\ 
Phase 3 & 2.42 & 0.005 & 1.08 & 1.92 & 3.20 \\ 
Phase 4 & 2.06 & 0.003 & 0.92 & 1.67 & 2.83 \\
\bottomrule
\end{tabular}
\label{tab:dura_stats}
\end{table*}

The data shows that predicting duration is challenging, with 75\% of the training data spanning up to 3 years and a maximum cap of 10 years.

Our final dataset comprised two distinct sets: a training set with 77,818 clinical trial records and a testing set with 36,786 records. We partitioned the data on January 1, 2019, using the data from trials that started before that date for training and validation and the data from trials that started after that date for testing. This approach ensures that the completion dates of the trials in the training and validation sets precede the start dates of the trials in the testing set.
We studied the general trial duration problem across different trial phases for many diseases, as shown in Table~\ref{tab:phase_distribution}. 

\begin{table}[htbp]
\centering
\caption{ Distribution of clinical trial phases. }
{%
\begin{tabular}{ccccc}
\toprule
Phases & 1 & 2 & 3 & 4  \\ \midrule
Number of Trials & 31,641 & 44,622 & 28,250 & 25,060 \\ 
\bottomrule
\end{tabular}
}
\label{tab:phase_distribution}
\end{table}

\subsection{Baseline Methods}
The Baseline Method is essential for contextualizing the performance of our proposed model, TrialDura, by comparing it with established machine-learning methodologies. To establish a benchmark, we chose multiple widely recognized models as baselines: 
Linear Regression (LR), Gradient Boosting Decision Tree (GBDT), Adaptive Boosting (AdaBoost), Random Forest (RF), and Multiple Layer Perceptron (MLP). 
GBDT, AdaBoost, RF, and MLP have been successfully applied for clinical trial outcome prediction~\citep{fu2022hint,fu2023automated,chen2024uncertainty}. 
We use MEAN as a reference, which uses the average duration as an estimation directly and offers a basic point of reference
We selected Linear Regression because of its simplicity and effectiveness in revealing linear relationships within the dataset, serving as a fundamental point of reference for performance evaluation. This provides an essential baseline that allows us to appreciate the incremental value added by more complex models. The Gradient Boosting Decision Tree (GBDT) was chosen for its proficiency in managing diverse and complicated datasets through the sequential correction of prediction errors. Its inclusion as a baseline underscores our commitment to assess our model against sophisticated state-of-the-art algorithms. Similarly, the Random Forest model was incorporated into our baseline comparison owing to its robustness and effectiveness in dealing with high-dimensional data. The risk of overfitting is reduced by aggregating the outcomes of numerous decision trees, thereby offering a reliable baseline for performance comparison.

By benchmarking TrialDura against these distinct and well-established models, we aim to demonstrate the specific advancements our model brings to the domain of clinical trial duration prediction. This comparison not only validates TrialDura's effectiveness but also situates its performance within the broader landscape of machine learning applications in clinical research, highlighting its contribution to enhancing predictive accuracy and decision-making in clinical trial management.

\subsection{Experimental Results}

\begin{table*}[htbp]
\centering
\caption{Summary of model performance (units in years). Results averaged over five runs; standard deviations were shown. Asterisk (*) denotes statistical significance (p-value $<$ 0.05). NA: not applicable.}
\begin{tabular}{ccccc}
\toprule
Model & MAE (years) ($\downarrow$) & RMSE (years) ($\downarrow$) & $R^2$ ($\uparrow$) & Pearson Corr. ($\uparrow$) \\ 
\midrule 
MEAN & 1.240 & 1.551 & -0.090 & NA \\ 
Linear Regression & 1.162 $\pm$ 0.005 & 1.511 $\pm$ 0.003 & 0.043 $\pm$ 0.002 & 0.275 $\pm$ 0.001 \\
GBDT & 1.148 $\pm$ 0.001 & 1.494 $\pm$ 0.002 & 0.080 $\pm$ 0.001 & 0.288 $\pm$ 0.001 \\
RF & 1.167 $\pm$ 0.001 & 1.502 $\pm$ 0.001 & 0.053 $\pm$ 0.001 & 0.232 $\pm$ 0.003 \\
XGBoost & 1.152 $\pm$ 0.004 & 1.497 $\pm$ 0.004 & 0.069 $\pm$ 0.001 & 0.279 $\pm$ 0.001 \\
AdaBoost & 1.200 $\pm$ 0.003 & 1.514 $\pm$ 0.002 & -0.258 $\pm$ 0.001 & 0.169 $\pm$ 0.002 \\
MLP & 1.144 $\pm$ 0.008 & 1.502 $\pm$ 0.011 & -0.749 $\pm$ 0.049 & 0.197 $\pm$ 0.000 \\
TrialDura & \textbf{1.044 $\pm$ 0.009}* & \textbf{1.390 $\pm$ 0.005}* & \textbf{0.191 $\pm$ 0.001}* & \textbf{0.463 $\pm$ 0.001}* \\
\bottomrule
\end{tabular}
\label{tab:model_performance}
\end{table*}

The performance of various predictive models for clinical trial duration is detailed in Table~\ref{tab:model_performance}. Metrics used for evaluation include Mean Absolute Error (MAE), Root Mean Squared Error (RMSE), the coefficient of determination ($R^2$), and the Pearson correlation coefficient. The MEAN model serves as a basic reference, underscoring the enhanced accuracy of more advanced models. For example, Linear Regression, although simplistic, achieves an MAE of 1.16 years and an RMSE of 1.511 years, demonstrating a basic yet effective capacity to model trial duration.

More sophisticated models such as Gradient Boosted Decision Trees (GBDT) and Random Forest (RF) show a further reduction in prediction errors, illustrating their capability in managing complex data structures. Notably, GBDT achieves an MAE of 1.14 years and an RMSE of 1.494 years. Models like XGBoost and Multi-Layer Perceptrons (MLP), while displaying similar MAE values, exhibit variations in other metrics, which may indicate differences in model sensitivity and the distribution of errors.

Our novel model, TrialDura, significantly outperforms all benchmarks, achieving the lowest MAE and RMSE, which are statistically significant improvements of 9\% and 7\% respectively over the best baseline model. This superior performance is corroborated by its $R^2$ and Pearson correlation scores, confirming its robust predictive capability.

We also show TrialDura's prediction performance for each of the four phases in Table~\ref{tab:phase_performance}, Results show the model's lowest MAE in Phases 1 and 4, highlighting its strengths and areas for improvement.

\begin{table*}[htbp]
\centering
\caption{Summary of model performance across different phases.}
\begin{tabular}{ccccc}
\toprule[1pt]
Phase & MAE (years) ($\downarrow$) & RMSE (years) ($\downarrow$) & $R^2$ ($\uparrow$) & Pearson Corr. ($\uparrow$) \\
\midrule 
All & 1.044 $\pm$ 0.009 & 1.390 $\pm$ 0.005 & 0.191 $\pm$ 0.001 & 0.463 $\pm$ 0.001 \\
1 & 0.836 $\pm$ 0.001 & 1.181 $\pm$ 0.000 & 0.342 $\pm$ 0.000 & 0.593 $\pm$ 0.000 \\
2 & 1.130 $\pm$ 0.001 & 1.459 $\pm$ 0.002 & 0.061 $\pm$ 0.002 & 0.334 $\pm$ 0.001 \\
3 & 1.194 $\pm$ 0.003 & 1.593 $\pm$ 0.002 & 0.159 $\pm$ 0.002 & 0.415 $\pm$ 0.002 \\
4 & 1.099 $\pm$ 0.001 & 1.402 $\pm$ 0.001 & -0.003 $\pm$ 0.002 & 0.228 $\pm$ 0.002 \\
\bottomrule[1pt]
\end{tabular}
\label{tab:phase_performance}
\end{table*}

\begin{figure*}[ht]
\centering
\includegraphics[width=\textwidth,keepaspectratio]{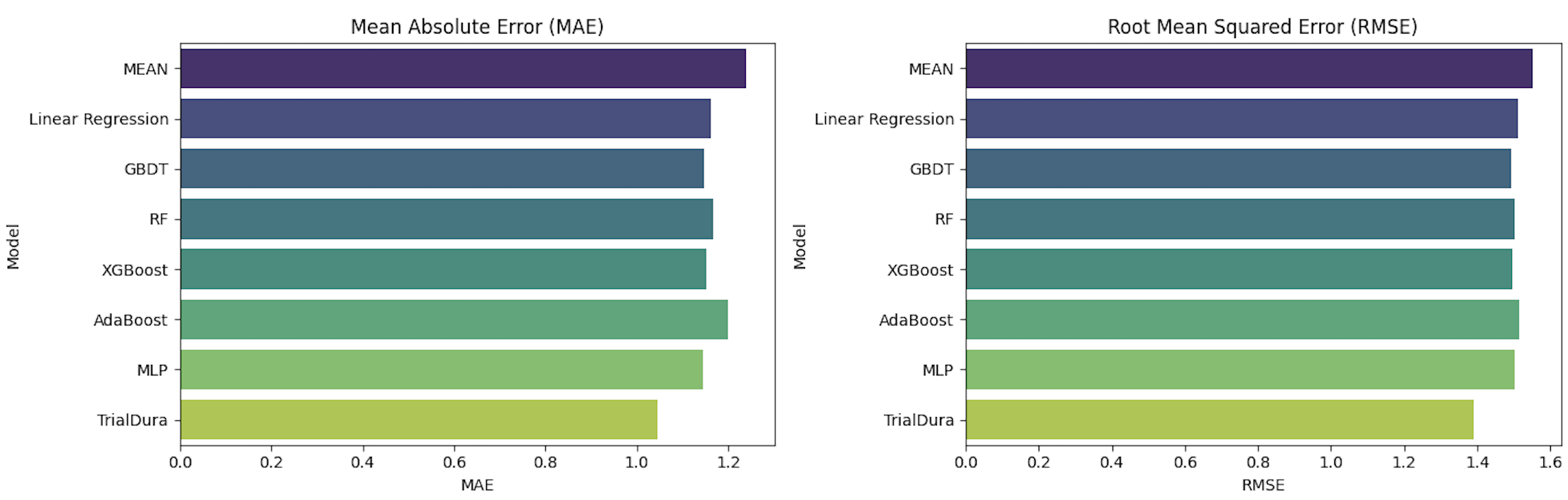}
\caption{Model performance comparison on prediction accuracy. }
\label{fig:barchart}
\end{figure*}

Figure~\ref{fig:barchart} visually summarizes these findings, illustrating the superior performance of TrialDura in both MAE and RMSE compared to the other models. The graph provides a clear visual representation of the comparative predictive accuracies, reinforcing the textual data presented in Table~\ref{tab:model_performance}.

In addition to quantitative metrics, qualitative analysis of these models suggests that the integration of domain-specific features into TrialDura, such as trial phase and therapeutic area, contributes significantly to its enhanced performance. Future work will focus on refining these features and exploring the incorporation of real-time data updates to further improve the model’s predictive accuracy and operational utility in clinical trial management.

\subsection{Ablation Study}
\label{sec:ablation}

This ablation study assesses the empirical impact of different model components on predicting clinical trial duration. We evaluate models separately trained for Phases 1 through 4 and compare them against a unified model, TrialDura, which integrates data across all phases.

The results, detailed in Table~\ref{tab:model_performance_phases_years}, indicate that the performance varies across phases. For example, Phase 1, tested on 9,816 data points, yielded a Mean Absolute Error (MAE) of approximately 1.12 years and a Root Mean Squared Error (RMSE) of about 1.42 years. The negative $R^2$ values and negligible Pearson correlation coefficients across all phases suggest challenges in phase-specific predictions. Conversely, the unified TrialDura model demonstrates superior performance with a significant positive $R^2$ and Pearson correlation, indicating a robust model that effectively captures the underlying patterns across different trial phases.

\begin{table*}[ht]
\centering
\caption{Ablation study: model performance across phases compared to the unified TrialDura model.}
\label{tab:model_performance_phases_years}
\begin{tabular}{lcccc}
\toprule
Phase & MAE (years) ($\downarrow$) & RMSE (years) ($\downarrow$) & $R^2$ ($\uparrow$) & Pearson Corr. ($\uparrow$) \\
\midrule
1 & $1.117 \pm 0.030$ & $1.424 \pm 0.011$ & $-0.146 \pm 0.017$ & $-0.002 \pm 0.002$ \\
2 & $1.227 \pm 0.038$ & $1.513 \pm 0.024$ & $-0.186 \pm 0.045$ & $0.005 \pm 0.006$ \\
3 & $1.327 \pm 0.010$ & $1.732 \pm 0.008$ & $-0.123 \pm 0.017$ & $0.005 \pm 0.004$ \\
4 & $1.125 \pm 0.010$ & $1.386 \pm 0.005$ & $-0.148 \pm 0.036$ & $-0.001 \pm 0.006$ \\ 
TrialDura & \textbf{1.044 $\pm$ 0.009} & \textbf{1.390 $\pm$ 0.005} & \textbf{0.191 $\pm$ 0.001} & \textbf{0.463 $\pm$ 0.001} \\ 
\bottomrule
\end{tabular}
\end{table*}

Additionally, we explored various methods for aggregating sentence embeddings into inclusion and exclusion criteria embeddings. Table~\ref{tab:sentence_aggregation} compares the efficacy of maximum pooling, mean pooling, and the CLS token method. The results are closely comparable, with mean pooling slightly outperforming the others by a narrow margin in $R^2$ value, suggesting that each method effectively captures essential features for the prediction task.

\begin{table*}[htbp]
\centering
\caption{Ablation study: comparison of aggregation methods for sentence representation.}
\label{tab:sentence_aggregation}
\begin{tabular}{lcccc}
\toprule
Method & MAE (years) ($\downarrow$) & RMSE (years) ($\downarrow$) & $R^2$ ($\uparrow$) & Pearson Corr. ($\uparrow$) \\
\midrule
Max Pooling & $1.048 \pm 0.001$ & $1.391 \pm 0.002$ & $0.189 \pm 0.005$ & $0.463 \pm 0.002$ \\
Mean Pooling & $1.048 \pm 0.001$ & $1.389 \pm 0.002$ & $0.191 \pm 0.001$ & $0.463 \pm 0.001$ \\
CLS Token & $1.048 \pm 0.001$ & $1.390 \pm 0.002$ & $0.191 \pm 0.002$ & $0.463 \pm 0.001$ \\ 
\bottomrule
\end{tabular}
\end{table*}

This systematic analysis underscores the importance of each component's contribution to the model's overall predictive performance and provides insights for future enhancements.

\subsection{Case Study}
\label{sec:case_study}

\begin{table*}[htbp]
\caption{Comparison of predicted and actual duration of clinical trials.}
\centering
\begin{tabular}{cccccccc}
\toprule
{NCTID} & {Phase} & {Disease} & {Drug} & {$y$ (Actual Years)} & {$\widehat{y}$ (Predicted Years)} \\
\midrule 
NCT03553810 & II & Hypertensive heart disease & Entresto & 5.00 & 4.64 \\ 
NCT03681093 & III & Asthma & Fevipiprant & 4.00 & 3.95 \\ 
NCT02198209 & IV & Type 2 diabetes & Sitagliptin & 2.37 & 2.11 \\ 
NCT03636373 & IV & Gout attack & Etanercept & 1.77 & 1.80 \\ 
NCT04126317 & II & Neovascular macular degeneration & Aflibercept & 2.51 & 1.92 \\ 
NCT04249882 & III & Depression & Naltrexone & 3.22 & 2.49 \\ 
NCT04050488 & II & Retinopathy of prematurity & Bevacizumab & 3.00 & 2.87 \\
NCT04388852 & I & Prostate carcinoma & Bendamustine & 4.03 & 4.04 \\ 
NCT04483778 & II & Neuroblastoma & Fludarabine & 4.99 & 5.00 \\ 
NCT04023669 & I & Brain tumor & Prexasertib & 3.71 & 3.73 \\ 
\bottomrule
\end{tabular}
\label{tab:case}
\end{table*}

Clinical trials are crucial for advancing medical science, providing essential data on the efficacy and safety of new treatments. Structured into four phases—Phase I (safety), Phase II (efficacy and side effects), Phase III (comparative effectiveness), and Phase IV (post-market surveillance)—each phase plays a vital role in the comprehensive assessment of new treatments. Accurate predictions of trial duration are essential for effective planning, resource allocation, and regulatory compliance, making this a key area for the application of machine learning.

\noindent\textbf{Examples of successful trials:}
Our model, TrialDura, was applied to a series of recent clinical trials, demonstrating its utility in accurately estimating trial duration. The trials, detailed in Table~\ref{tab:case}, include various drugs and conditions, showing the model's broad applicability. For instance, the trial for Entresto, a heart failure drug developed by Novartis, involved 4,822 patients and was predicted to last 4.9 years, closely matching the actual duration of five years. Despite the trial not meeting its primary endpoints, the accurate prediction underscores the financial implications and potential for improved planning and resource allocation. Similarly, the phase III trial for Fevipiprant, aimed at treating asthma, lasted four years, with our model's prediction being just under 3.95 years. Although the trial did not meet its anticipated goals, the precision of our predictions highlights the model's capability to aid the financial forecasting and strategic planning of clinical trials.

\begin{figure}[htbp]
\centering

\begin{tcolorbox}[colback=gray!5, sharp corners, boxrule=0.5pt, boxsep=1mm, left=1mm, right=1mm, top=1mm, bottom=1mm, center title]
\textbf{\fcolorbox{black}{yellow!50}{0.85}} \colorbox{red!60}{\textbf{Increased}} \colorbox{red!10}{left} \colorbox{red!60}{\textbf{ventricular}} \colorbox{red!60}{\textbf{mass}} \colorbox{red!10}{on} \colorbox{red!30}{cardiovascular} \colorbox{red!60}{\textbf{magnetic}} \colorbox{red!60}{\textbf{resonance}} 
\end{tcolorbox}

\begin{tcolorbox}[colback=gray!5, sharp corners, boxrule=0.5pt, boxsep=1mm, left=1mm, right=1mm, top=1mm, bottom=1mm, center title]
\textbf{\fcolorbox{black}{yellow!20}{0.55}} \colorbox{red!30}{Known} \colorbox{red!60}{\textbf{secondary}} \colorbox{red!60}{\textbf{causes}} \colorbox{red!10}{of} \colorbox{red!30}{hypertension}
\end{tcolorbox}

\begin{tcolorbox}[colback=gray!5, sharp corners, boxrule=0.5pt, boxsep=1mm, left=1mm, right=1mm, top=1mm, bottom=1mm, center title]
\textbf{\fcolorbox{black}{yellow!20}{0.50}} \colorbox{red!30}{Previous} \colorbox{red!60}{\textbf{intolerance}} \colorbox{red!10}{to} \colorbox{red!60}{\textbf{angiotensin}} \colorbox{red!60}{\textbf{receptor}} \colorbox{red!60}{\textbf{blockers}}
\end{tcolorbox}

\caption{
Visualization of text segments in the TrialDura model's output, illustrating Shapley values derived from Clinical Trials (NCT03553810). Shapley values correspond to attention weights, with darker colors indicating higher weights. The number and color at the beginning of each sentence represent the attention weight for the entire sentence.
}
\label{fig:sentence_importance}
\end{figure}

\subsection{Interpretability Analysis}

The visualization of word and sentence importance in the TrialDura model's eligibility criteria can be clearly demonstrated using Shapley values, 
which are depicted in Figure~\ref{fig:sentence_importance}. 
The Shapley value~\citep{winter2002shapley}, from game theory, fairly distributes total gains or costs among players based on their contributions. In machine learning, Shapley values measure each feature's impact on a model's prediction. 
These values show the incremental contribution of each word and sentence in the text as it transitions from a fully masked to an unmasked state, 
highlighting the impact of each component on the model's final output. 
This process is quantified starting from a base logit value obtained when all the input text is masked, 
moving to specific word and sentence contributions as each is revealed. 

In our model, more attention is paid to words and sentences with higher Shapley values, as indicated by darker coloration in the visual representation. 
This method provides a deep understanding of how different terms in the eligibility criteria affect a model's predictions. 
By visualizing these values, we can discern which words or phrases are deemed most critical by the model in determining trial duration, 
thereby offering insights into the model's decision-making processes and helping identify areas for further model refinement or feature engineering. 
This approach not only enhances interpretability but also aids in validating the model's focus on relevant aspects of the input data, ensuring that the model's attention aligns with logical and clinical expectations from the trial data.

\section{Discussion}

This study marks a notable progression in leveraging machine learning to forecast the duration of clinical trials, integrating multimodal data through the advanced architecture of TrialDura. This model, equipped with a hierarchical attention mechanism, adeptly handles complex datasets involving drug characteristics, disease categories, and intricate eligibility criteria. Such a comprehensive approach allows for a detailed analysis of factors influencing trial timelines, providing valuable insights for clinical trial management.

\paragraph{Technical Implications}
Technically, TrialDura introduces a novel use of hierarchical attention mechanisms, specifically designed to address the multifaceted nature of clinical trial data. This approach enhances interpretability, a key feature in clinical applications where understanding the rationale behind model predictions is as critical as the predictions themselves. Additionally, the incorporation of Bio-BERT for processing biomedical text significantly improves the model's ability to interpret complex medical documentation. This specificity could be beneficial in other healthcare applications where accurate interpretation of technical language is essential.

\paragraph{Clinical Implications}
From a clinical perspective, the predictive accuracy of TrialDura can significantly improve the management of clinical trials. By providing reliable estimates of trial durations, the model aids in optimizing resource distribution, refining recruitment strategies, and reducing overall costs and development times for new medical treatments. These capabilities are crucial for enhancing the efficiency of clinical trials, especially in an environment where delays are both common and costly.

\paragraph{Limitations}
Despite its advancements, TrialDura faces challenges in achieving precise predictions, evidenced by a mean absolute error (MAE) of over one year. This inaccuracy stems from the inherent complexity of clinical trial data, the limited availability of robust training datasets, and the model's current inability to adapt to ongoing changes in regulatory and legal frameworks. Future improvements should focus on refining data processing techniques, enhancing the model's sensitivity and specificity, and increasing its adaptability to continuously evolving conditions.

In summary, TrialDura represents a significant advancement in applying machine learning to clinical trials, offering not only improved predictive capabilities but also deeper insights into the dynamics influencing trial durations. This development underlines the growing intersection of machine learning and healthcare, promoting a more data-driven approach to clinical trial management that could lead to more streamlined and effective healthcare solutions.

\bibliographystyle{ACM-Reference-Format}
\bibliography{reference}

\appendix

\section{Appendix}

\subsection{Evaluation Metrics} 

To evaluate the prediction performance of our TrialDura model, we employed the Mean Absolute Error (MAE) and Root Mean Square Error (RMSE), defined by the following equations: 
\begin{equation}
\text{MAE} = \frac{1}{n}\sum_{i=1}^{n} |y_i - \widehat{y}_i|, 
\end{equation}
\begin{equation}
\text{RMSE} = \sqrt{\frac{1}{n}\sum_{i=1}^{n}(y_i - \widehat{y}_i)^2},
\end{equation}
where $y_i$ and $\widehat{y}_i$ represent the true and predicted duration for the $i^{th}$ clinical trial, respectively, and $n$ is the total number of trials in the test dataset.

\noindent\textbf{R-squared ($R^2$) score} is defined as the proportion of the variation in the dependent variable that is predictable from the independent variable(s). It is also known as the coefficient of determination in statistics. 
 Suppose we have $N$ continuous ground truth $y_1, \cdots, y_N$ and $N$ corresponding predictions $\widehat{y_1}, \cdots, \widehat{y_N}$. The difference $y_i - \widehat{y_i}$ is called residual. Then, we defined the residual sum of squares as 
 \begin{equation}
 \text{SS}_{\text{res}} = \frac{1}{N} \sum_{i=1}^{N} (y_i - \widehat{y_i})^2,     
 \end{equation}
 and define the total sum of squares as
 \begin{equation}
 \text{SS}_{\text{total}} = \frac{1}{N} \sum_{i=1}^N (y_i - \bar{y})^2,     
 \end{equation}
 where $\bar{y} = \frac{1}{N} y_i$ is the mean of the ground truth. 
 Then $R^2$ score is defined as 
 \begin{equation}
 R^2 = 1 - \frac{\text{SS}_{\text{res}}}{\text{SS}_{\text{total}}}.     
 \end{equation}
Higher $R^2$ scores indicate better performance. 
In a perfect prediction model, the prediction exactly matches the ground truth, then $\text{SS}_{\text{res}} = 0$ and $R^2 = 1$. 
A weak predictor that always predicts $\bar{y}$ would cause $R^2 = 0$. 
Some predictors would even lead to a negative $R^2$ score.

\noindent\textbf{Pearson Correlation} (PC) is defined as the covariance of the prediction and the ground truth divided by the product of their standard deviations. 
 For two random variables $x$ and $y$, Pearson Correlation is formally defined as 
 \begin{equation}
 \text{PC} = \frac{\mathbb{E}[(x - \mu_{x}) ({y} - \mu_{y})]}{\sigma_{x} \sigma_{y}}, 
 \end{equation}
 In the regression task, suppose there are $N$ data points in the test set, $y_i$ is the ground truth of the $i$-th data sample, $\widehat{y}_i$ is the prediction for $i$-th data, Pearson Correlation becomes 
 \begin{equation}
 \text{PC} = \frac{ \sum_{i=1}^{N} \big( (y_i - \mu_y) (\widehat{y}_i - \mu_{\widehat{y}}) \big) }{ \sigma_{y} \sigma_{\widehat{y}} }, 
 \end{equation}
 where $\mu_{y} = \frac{1}{N}\sum_{j=1}^{N} y_{j}$ and $\mu_{\widehat{y}} = \frac{1}{N}\sum_{j=1}^{N} \widehat{y}_{j}$ are mean of ground truth and prediction, respectively. 
 $\sigma_{y} = \sum_{i=1}^N (y_i - \frac{1}{N}\sum_{j=1}^{N} y_{j})^2$ and $\sigma_{\widehat{y}} = \sum_{i=1}^N (\widehat{y}_i - \frac{1}{N}\sum_{j=1}^{N} \widehat{y}_{j})^2$ are the standard deviations of ground truth and prediction, respectively. 
 The value ranges from -1 to 1. 
 A higher Pearson correlation value indicates better performance.

Also, we conduct hypothesis testing by evaluating the p-value to showcase the statistical significance of our method over the best baseline results. If the p-value is smaller than 0.05, we reject the null hypothesis and claim our method significantly outperforms the best baseline method.

\subsection{Implementation Details and Hyperparameter Tuning} 

We employed Bayesian Optimization with TPE (Tree-structured Parzen Estimators) for hyperparameter tuning using the Optuna~\citep{10.1145/3292500.3330701} library. In our configuration, we set the number of epochs to 1500 and allowed the number of attention heads to vary among [2, 3, 4]. The number of hidden units was set to (2*768) for the first linear layer, 768 units for the second linear layer, and 768 units for the transformer feed-forward layer. We maintained the dropout ratio at 0.1 and set the learning rate at 0.001 to optimize our model's performance efficiently.
The model was trained for 200 epochs at a learning rate of 0.001 to optimize the network weights and biases to effectively reduce the loss function.

\end{document}